\documentclass[letterpaper, 10 pt, conference]{ieeeconf}  

\IEEEoverridecommandlockouts                              

\usepackage{graphics} 
\usepackage{epsfig} 
\usepackage{mathptmx} 
\usepackage{times} 
\usepackage{multirow}
\usepackage{amsmath} 
\usepackage{amssymb}  
\usepackage{algorithmic}
\usepackage{algorithm}
\usepackage{balance}
\usepackage{soul,xcolor}
\usepackage[hidelinks, implicit=false]{hyperref}
\let\labelindent\relax
\usepackage{enumitem}
\usepackage{scalerel}
\setitemize{leftmargin=*}

\usepackage{dsfont}
\usepackage{array}
\usepackage{threeparttable}
\usepackage[noadjust]{cite}
\newcolumntype{P}[1]{>{\centering\arraybackslash}p{#1}}

\title{\LARGE \bf
Learning Agile Locomotion on Risky Terrains
}

\author{Chong Zhang$^{1}$, Nikita Rudin$^{1}$, David Hoeller$^{1}$, Marco Hutter$^{1}$
\thanks{$^{1}$ Authors are with the Robotic Systems Lab, ETH Zurich, 8092 Zurich, Switzerland. Emails: {\tt\small \{chozhang, rudinn, dhoeller, mahutter\}@ethz.ch}}%
}

\begin{document}

\maketitle
\thispagestyle{empty}
\pagestyle{empty}

\begin{abstract}
Quadruped robots have shown remarkable mobility on various terrains through reinforcement learning. Yet, in the presence of sparse footholds and risky terrains such as stepping stones and balance beams, which require precise foot placement to avoid falls, model-based approaches are often used. In this paper, we show that end-to-end reinforcement learning can also enable the robot to traverse risky terrains with dynamic motions. To this end, our approach involves training a generalist policy for agile locomotion on disorderly and sparse stepping stones before transferring its reusable knowledge to various more challenging terrains by finetuning specialist policies from it. Given that the robot needs to rapidly adapt its velocity on these terrains, we formulate the task as a navigation task instead of the commonly used velocity tracking which constrains the robot's behavior and propose an exploration strategy to overcome sparse rewards and achieve high robustness. We validate our proposed method through simulation and real-world experiments on an ANYmal-D robot achieving peak forward velocity of $\ge 2.5$ m/s on sparse stepping stones and narrow balance beams. Video: youtu.be/Z5X0J8OH6z4
\end{abstract}

\section{INTRODUCTION}

Quadrupedal robots require perceptive locomotion abilities on rough and complex terrains, including risky terrains with sparse footholds such as stepping stones and balance beams. Existing works typically employ model-based controllers \cite{winkler2018gait,jenelten2022tamols,fahmi2022vital,grandia2023perceptive,melon2021receding,jenelten2020perceptive} which perform well in simulation or under laboratory conditions. However, they are empirically difficult to deploy in the wild due to model mismatch and unexpected slippage \cite{lee2020learning,jenelten2019dynamic}, and typically require dedicated design (e.g. motion references \cite{zhou2022momentum} and handcrafted heuristics such as terrain roughness and pose safety \cite{fahmi2022vital}) to achieve specified motions. Furthermore, the dynamics model needs to be simplified especially for the limbs \cite{jenelten2022tamols, grandia2023perceptive}, and replanning is often slow due to the heavy computational burden \cite{fahmi2022vital,melon2021receding}.

On the other hand, model-free reinforcement learning (RL)-based methods demonstrate excellent robustness in the wild via domain randomization and massive training data \cite{miki2022learning,tranzatto2022cerberus}. However, no existing works have used end-to-end RL to handle the sparse footholds on highly risky terrains, where precise motions are necessary to avoid failures, and exploration is difficult due to the numerous ways to fail. First attempts to solve such locomotion challenges using end-to-end learning show promising results, but still have limited terrain complexity \cite{agarwallegged} or need to exactly know which stones to step on \cite{xie2020allsteps}. Another work trains decoupled policies for foothold planning and execution \cite{tsounis2020deepgait}, but it cannot achieve dynamic locomotion on risky terrains, and decoupled planning and control can inherently limit the range of behaviors that can be expressed \cite{song2023reaching}.

Some existing works combine RL and model-based controllers. For example, \cite{gangapurwala2022rloc} uses RL to generate desired foot positions which are then tracked by a whole-body motion controller, as well as corrective joint torques to achieve domain adaptative tracking. Another work uses RL to generate the desired acceleration of the base and tracks it with model-based controllers \cite{xie2023glide}. All of these approaches, however, demonstrate limited agility with velocity $\le 0.5 \rm m/s$. Another recent work \cite{jenelten2023dtc} opens a new venue by tracking trajectories from a model-based planner with an RL policy and demonstrates remarkable performance, but the motions are constrained by the planner and cannot recover from failures.



\begin{figure}[t]
  \centering
    \includegraphics[width=85mm]{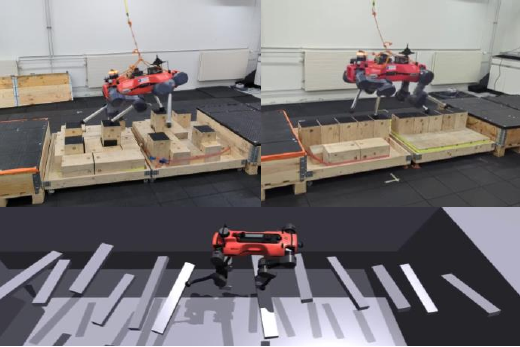}
    \caption{
       The ANYmal-D robot executing the learned RL policies on stepping stones, balance beams with gaps, and stepping beams. \textbf{Left top:} the stepping stones were $20$ cm wide taking up only $\sim 10\%$ of the whole area, and the height difference between high and low stones was $12$ cm. The robot reached the peak forward velocity of $2.7$ m/s when traversing the stones from left to right. \textbf{Right top:} the terrain consisted of two beams of $100$ cm $\times$ $20$ cm and three gaps. The peak forward velocity of the robot was $2.5$ m/s from right to left. \textbf{Bottom:} The widths of stepping beams varied in $[12$ cm$, 17$ cm$]$, the horizontal distances between two beams varied in $[30$ cm$, 60$ cm$]$, and the vertical distances varied in $[0$ cm$, 20$ cm$]$. For practical reasons, we only tested the policy in simulation, and the peak forward velocity of the robot reached $1.5$ m/s from left to right.
    }
    \label{fig:fig1cover}
  \vspace{-6mm}
\end{figure}

In this paper, we explore how dynamic locomotion of a quadrupedal robot can be achieved on highly risky terrains through end-to-end RL. We first train a generalist policy to learn reusable sensorimotor skills on sparse and disorderly stepping stones. Then, we finetune multiple specialist policies from the generalist policy on various more challenging terrains, and successfully transfer two of the specialist policies to real-world stepping stones and narrow beams. We achieve highly agile motions with peak forward velocity $\geq 2.5$ m/s, as shown in Fig.~\ref{fig:fig1cover}.

\begin{figure*}[t]
   \centering
   \includegraphics[width=177mm]{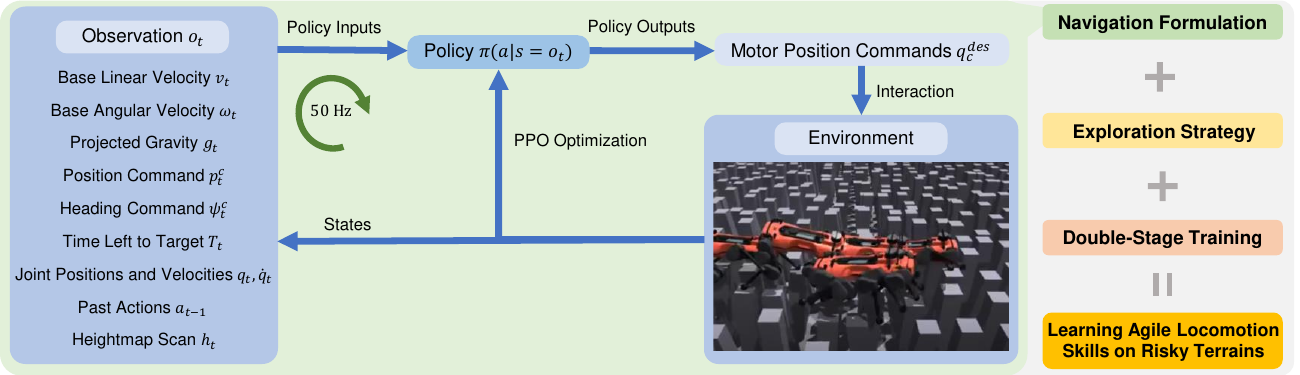}
   \caption{An overview of our methodology. The learning system is illustrated on the left. Different components are listed on the right.}
   \label{fig:policy_obs}
   \vspace{-4mm}
\end{figure*}

Traditional learning methods based on the commonly used velocity tracking formulation (e.g., the one used in \cite{rudin2022learning} and \cite{agarwallegged}) can not achieve these challenging tasks. It is intractable to assign appropriate velocity commands to the robots on risky terrains especially when they need to rapidly adapt their velocities, and the feasible solutions are quite rare and hard to explore compared to nonrisky terrains. Hence, the key idea in this paper is to formulate the task as a navigation problem as in \cite{rudin2022advanced}, though the risky environments and the sparse rewards make the exploration even harder.

To overcome the hard exploration in the navigation formulation, we improve the curriculum learning design in \cite{rudin2022advanced}, and incorporate curiosity-driven intrinsic rewards~\cite{RND}. We also employ a symmetry-based data augmentation from \cite{hoeller2023anymal}, utilizing the left-right and front-back symmetry of our robot to improve the data efficiency. These improvements result in an exploration strategy that can learn a robust generalist policy. 

Still, we find such an exploration strategy above is not enough to learn specialist policies from scratch on the terrains in Fig.~\ref{fig:fig1cover}. Besides, we want the learned sensorimotor skills to be reusable for new terrains. To this end, we divide the training procedure into two stages, where the first stage is to learn a common generalist policy on sparse and disorderly stepping stones, and the second stage is to finetune different specialist policies on different terrains from this generalist policy. By doing so, the generalist policy can learn some basic motion skills that are reusable on different terrains, and the specialist policies can benefit from the pre-training of the first stage. This generalist-specialist training procedure enables learning of agile locomotion skills which are then transferred to a real robot on risky terrains.

It is also worth mentioning that we do not focus on perception in this paper and therefore use a ground truth map and a motion capture system for state estimation during real-world deployment. That said, it is possible to use onboard sensors with the techniques proposed in~\cite{hoeller2022neural}.

In summary, our contributions include:
\begin{enumerate}
    \item The extension of the navigation formulation in \cite{rudin2022advanced} to challenging risky terrains;
    \item An exploration strategy consisting of a curriculum, intrinsic rewards, and a symmetry-based data augmentation;
    \item The methodology of training a generalist policy and finetuning specialist policies under hard exploration;
    \item Successful sim-to-real transfer of the learned agile motions on challenging stepping stones and balance beams.
\end{enumerate}

\section{Methodology}

All of the implementations in this paper are based on the PPO algorithm~\cite{schulman2017proximal} with massively parallel simulation~\cite{rudin2022learning} enabled by the GPU-based simulator Isaac Gym~\cite{makoviychuk2021isaac}. An overview is presented in Fig.~\ref{fig:policy_obs}.

\subsection{Formulating Locomotion as Navigation}

In this paper, we formulate the problem as a navigation task: the robot should get close to an assigned target on the terrain to get high task rewards. The navigation formulation has shown great potential when velocity commands cannot be easily assigned, e.g., when the robot walks in a maze \cite{chane2021goal} or needs to first recover from a fall and then walk~\cite{yang2020multi}. In \cite{rudin2022advanced}, advanced locomotion skills like jumping over large gaps and climbing high boxes are learned with the navigation formulation, where velocity-based tracking fails to achieve the desired behaviors. 

In this work, considering that the robot should adapt its velocity to the terrain and inappropriate velocity commands can be misguiding, we also use the navigation formulation. We deploy the system illustrated in Fig. \ref{fig:policy_obs} based on \cite{rudin2022advanced}, and assign position and heading commands to the robots. Tracking these commands dominates the rewards as follows:
\begin{equation}
r_{\text{task}}= \begin{cases}\frac{c_{\text{task}}}{T_r} \cdot \frac{1}{1+\left\|{\chi_t}-{\chi_t}^*\right\|^2}, & \text { if } t>T-T_r \\ 0, & \text { otherwise, }\end{cases}
\end{equation}
where $\chi$ and $\chi^*$ are the current and target positions (or headings) of the base, $t$ is the current time, $T$ is the maximum episode length, and $T_r$ defines the duration of the reward. By doing so, the policy is free to modulate the movement of the robot as long as it reaches the target towards the end of the episode.

\subsection{Exploration Strategy}

Our exploration strategy improves the prior work~\cite{rudin2022advanced} in three aspects: the curriculum learning design, the intrinsic rewards, and the data augmentation. The proposed techniques can empirically train a more robust policy under sparse rewards, and are experimentally analyzed in the ablation studies in Sec. \ref{subsec:ablation}.

\subsubsection{Curriculum Design}

Curriculum learning is a common technique for learning locomotion skills on challenging terrains \cite{lee2020learning,rudin2022learning,miki2022learning,rudin2022advanced}. In Isaac Gym, a typical implementation of curriculum learning is to generate all terrains of different difficulty levels before training, and distribute the robots on different terrains \cite{rudin2022learning,rudin2022advanced}. Based on this, the curriculum design in \cite{rudin2022advanced} is formulated as: 1) go to a higher difficulty level if the policy performs well, 2) go to a lower difficulty level if the policy performs poorly, and 3) go to a random difficulty level if the policy performs well on the highest difficulty level. 

However, we find that such a curriculum suffers from sparse rewards and difficulty gaps between levels due to the highly challenging and risky terrains. A policy performing well on a low-difficulty terrain can hardly work on a high-difficulty terrain, leading to a large reward difference that can be incorrectly attributed by the difficulty-agnostic optimizer to different actions, thereby misguiding the optimization. To alleviate this problem, we propose to relax the curriculum update mechanism: the robot only gets demoted when the policy cannot outperform random actions in the remaining distance to the target. By doing so, the robot can keep interacting with the same level until it overcomes it. 

\subsubsection{Intrinsic Curiosity Rewards}

On top of the task rewards which incentivize the robot to reach the target, we also use curiosity-based intrinsic rewards that encourage the robot to explore novel state-action pairs during training. Among many variants \cite{RND,pathak2017curiosity,burda2018largescale}, we use the random network distillation (RND) \cite{RND} with the version in \cite{sun2022optimistic} due to its simple implementation and competitive performance against other variants. Specifically, the curiosity reward is defined as
\begin{equation}
r_{\text{curio}}(s, a)=c_{\text{curio}} \cdot\left|M_1(s, a)-M_2(s, a)\right|,
\end{equation}
where $M_1$ and $M_2$ are two randomly initialized neural networks taking state-action $(s, a)$ pairs as inputs, and $M_1$ can be trained while $M_2$ is frozen. By fitting $M_1$ towards the outputs of $M_2$ with visited state-action pairs, the curiosity rewards will be high for unfamiliar pairs but decay to $0$ for familiar ones. 

It is worth mentioning that while curiosity rewards are commonly leveraged to improve data efficiency, we do not witness a significant change on the learning curves. Instead, we empirically find that it can improve the robustness of the learned policy, as is discussed in Sec. \ref{subsec:ablation}. This can potentially be due to an increased coverage of the state space, as suggested by \cite{zhang2022accessibility} where a larger state-space coverage can improve the robustness of some learned locomotion skills.

\subsubsection{Symmetry-Based Data Augmentation}


To improve the data efficiency and reduce base inclination during locomotion, we utilize the symmetric design of our robot to augment sampled data as proposed in \cite{hoeller2023anymal}, which is an adaptation of the duplication method in \cite{abdolhosseini2019learning}. To be specific, one state-action pair can be augmented to four pairs with left-right and front-back symmetry. The augmented state-action pairs are assigned the same target values, advantages, and discounted returns as the original pairs. In this way, the actor and critic networks are enforced to learn symmetric actions and values.

\subsection{Double-Stage Generalist-Specialist Training}

In this paper, we achieve agile locomotion on diverse challenging terrains with sparse footholds as shown in Fig. \ref{fig:fig1cover}. However, even with all the techniques presented above, these motions cannot be learned from scratch due to the extremely sparse rewards. To address this, we first train a generalist policy on relatively simple stepping stones (see Sec. \ref{subsec:terrains}). Then, the learned policy and its critic network are finetuned on new terrains, enabling data-efficient learning of agile motions, where the generalist policy can serve as a pre-trained model.

Besides the data efficiency, the generalist policy is shared for different specialist policies, which means that the learned sensorimotor skills are reusable. Therefore, the costly generalist training stage only happens once, and only data-efficient finetuning is needed for new terrains.

\section{Experimental Setup}

We create 4 different types of terrains in Isaac Gym, each with 10 difficulty levels. Among them, "Stones-Everywhere" is used for generalist policy training and benchmarking, while the other three ("Stones-2Rows", "Balance-Beams", and "Stepping-Beams") are used for finetuning. The terrains are described in Sec. \ref{subsec:terrains}. Sec. \ref{subsec:rewards} and Sec. \ref{subsec:dr} present the specific rewards and task-related randomizations of each terrain type respectively.

\subsection{Terrains}
\label{subsec:terrains}

\begin{figure}[t]
   \centering
   \includegraphics[width=86mm]{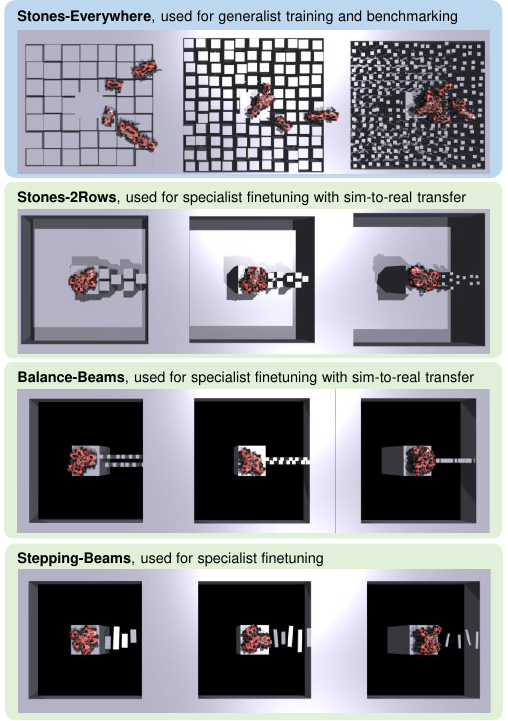}
   \caption{Different types of terrains with difficulty levels 0, 5, and 9 from left to right. One terrain type can have different subtypes that stress different features, as is explained in Sec. \ref{subsec:terrains}.}
   \label{fig:terrains}
   \vspace{-6mm}
\end{figure}

The four terrain types are displayed in Fig. \ref{fig:terrains} and described in the following paragraphs. Since the "Stones-Everywhere" type serves to benchmark the performance of different learning methods, we present it in detail, while other terrain types are introduced briefly.

\subsubsection{Stones-Everywhere}

This is the terrain type used to train the generalist policy and benchmark the ablation studies. Each stone is assigned a grid square and has a random height with a random X-Y plane shift from the square center. Highly disorderly and sparse stepping stones can be generated in this way. The parameters for different levels include the stone size $w_{\text{stone}}$, the stone gap $w_{\text{gap}}$, the maximum shift $s_{\text{max}}$, and the maximum height $h_{\text{max}}$. The stone size is the width of the square stones, and the stone gap is the grid size minus the stone size, i.e., the size of gap between two stones with zero shift. Both the X shift and the Y shift are uniformly sampled from [-$s_{\text{max}}$, $s_{\text{max}}$], and the stone height is uniformly sampled from [-$h_{\text{max}}$, $h_{\text{max}}$]. 

The parameters are presented in Table \ref{tab:ss_para}, together with the sparsity representing the ratio of the non-steppable area over the whole terrain. 


\subsubsection{Stones-2Rows} Terrains of this type have 1, 2 or 3 rows of stepping stones which match what we can build in the lab. By doing so we can achieve a policy that does not overfit to a specific gait pattern.

\subsubsection{Balance-Beams} 
\begin{figure}[!t]
  \centering
    \includegraphics[width=40mm]{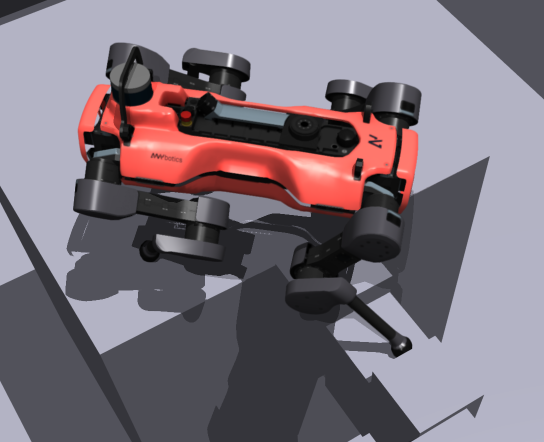}
    \caption{
       The learned infeasible crawling motions on balance beams due to inappropriate terrain curriculum design. The terrain here for display is far less challenging than the terrains in the real world, but the robot cannot traverse it. When the beam is narrow, the robot cannot find its next step on the beam.
    }
    \label{fig:crawl}
  \vspace{-6mm}
\end{figure}

Terrains of this type have stepping stones evolving from two rows merging into one as the level goes up. Random gaps are added to the terrains as well. Such evolving stones can guide the policy to learn catwalk-like motions. Beams with gradually smaller widths can be a more intuitive choice, but this can lead to infeasible crawling motions in Fig. \ref{fig:crawl}, a local minimum that works only on wide beams.

\subsubsection{Stepping-Beams} Terrains of this type have a sequence of beams to step on, and the beams are more narrow and randomized on higher levels. This terrain type is hard to build in the lab so we only tried it in simulation to show that finetuning can work on a wide range of different terrains.

\subsection{Rewards}
\label{subsec:rewards}

The same reward function can work on all terrains in simulation, yet slight terrain-dependant modifications of are necessary for elegant motions in the real world. We present the rewards in Table \ref{tab:reward} in the Appendix, and the changes in weights for special tasks are annotated. For each task on its terrain, the reward function is the sum of all terms.

Among these rewards, we have position tracking and heading tracking for task performance. We use penalties for early termination and collision. We also regularize the joint velocities, the acceleration of the base and feet, the action rate, the torques, and the contact forces. We guide exploration with the "Don't wait" and "Move in Direction" rewards as in ~\cite{rudin2022advanced}, and provide a "Stand Still" rewards to make the robot stand still after reaching the target.

We regularize the aggressive motions on "Stones-2Rows" and "Balance-Beams", and further regularize the torques on "Balance-Beams". We also add a "Stand Pose" reward to "Stones-2Rows" and "Balance-Beams" to prepare the robot for further repetitive tests.

\begin{table}[!t]
\caption{Parameters of "Stones-Everywhere" terrains}
\label{tab:ss_para}
\resizebox{86mm}{!}{%
\begin{threeparttable}
\begin{tabular}{P{1.2cm}P{1.4cm}P{1.4cm}P{1.4cm}P{1.6cm}P{1.2cm}}
\hline
\textbf{Difficulty Level} & \textbf{Stone Size $w_{\text{stone}}$ (cm)} & \textbf{Stone Gap $w_{\text{gap}}$ (cm)} & \textbf{Max. Shift $s_{\text{max}}$ (cm)} & \textbf{Max. Height $h_{\text{max}}$ (cm)}  & \textbf{Sparsity (\%)}\\ \hline
0 & 92 & 8  & 3.6  & 1 & 15.4\\ 
$\cdots$&$\cdots$&$\cdots$&$\cdots$&$\cdots$&\\
5 & 52 & 18 & 8.2  & 6 & 44.8\\
$\cdots$&$\cdots$&$\cdots$&$\cdots$&$\cdots$&\\
9 & 20 & 26 & 11.8 & 10& 81.1\\
\hline
\end{tabular}%
\begin{tablenotes}\footnotesize
\item[*] All parameters except the calculated sparsity are linear to the levels so we only list the levels 0, 5, and 9. 
\item[**] For reference, \cite{agarwallegged} built the stepping stones of $\sim 65 \%$ sparsity without random heights and significant shifts.
\end{tablenotes}
\end{threeparttable}
}
\vspace{-2mm}
\end{table}

\subsection{Domain Randomization}
\label{subsec:dr}

We use domain randomization to improve robustness and facilitate sim-to-real transfer \cite{tobin2017domain, xie2021dynamics}. Task-related randomization settings are presented in Table \ref{tab:dr} in the Appendix. We slightly extend the episode length to slow down the aggressive motions on "Stones-2Rows" and "Balance-Beams". The center of mass (CoM) is heavily randomized for "Balance-Beams" to reduce the dangerous shaking of the base on the beams.

\section{Results in Simulation}

\begin{figure*}[t!]
   \centering
   \includegraphics[width=175mm, trim = 0in 2.65in 0in 2.65in]{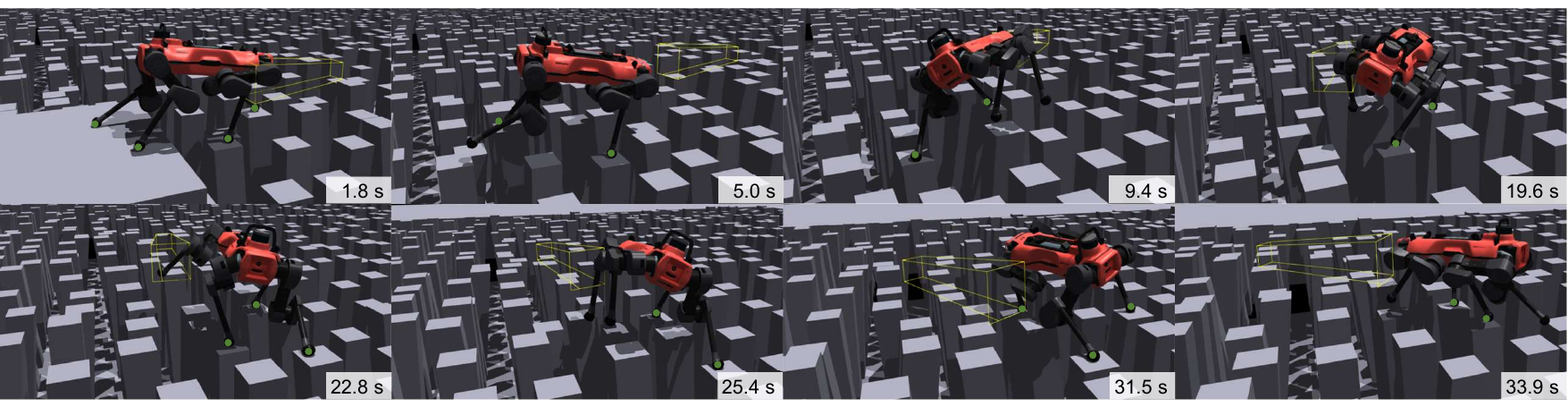}
   \caption{The robot walking around on the hardest "Stones-Everywhere" terrain in simulation. The yellow boxes indicate a series of target positions and headings we provided to the robot to achieve omnidirectional locomotion. The robot had different gait patterns under different situations, e.g., the 3-contact walk or 2-contact trot gaits when easy to find safe footholds ahead, and the 2-contact pace gaits otherwise. Feet in contact with the terrain are marked.
   }
   \label{fig:turning}
   \vspace{-4mm}
\end{figure*}

\begin{figure}[t!]
  \centering
    \includegraphics[width=86mm, trim = 0.cm 0.86in 0cm 0.82in]{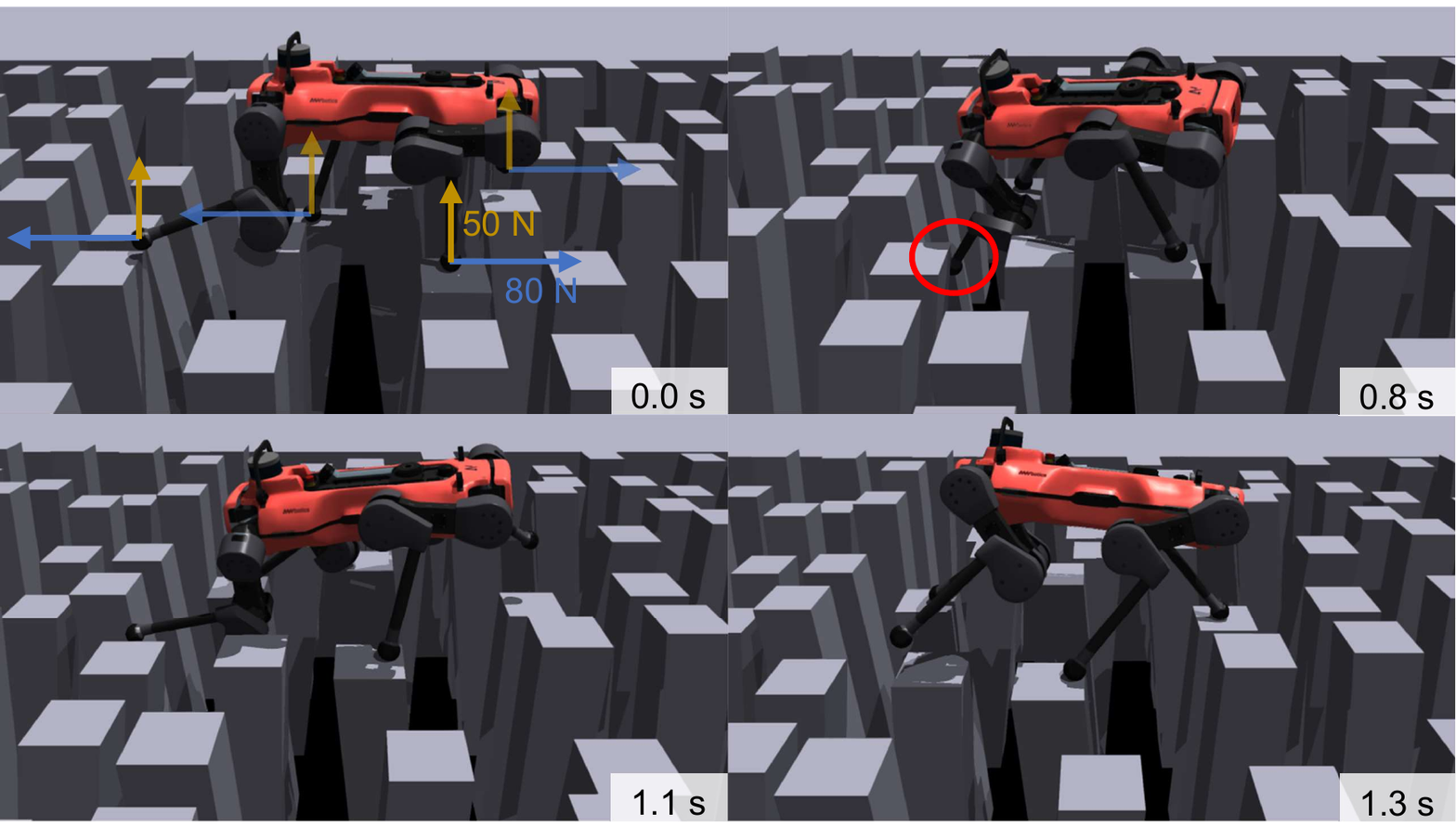}
    \caption{
       The robot could dynamically recover from missed steps due to external force disturbances while walking on stepping stones in simulation. \textbf{Left top:} the robot received external forces on its four feet lasting 0.4 s when traversing the stones from left to right. \textbf{Right top:} the robot missed a step due to the external forces. \textbf{Left bottom:} the robot recovered from the missed step by pulling itself upwards and forwards with the legs pressing the stones. \textbf{Right bottom:} the robot continued to walk after the missed step.
    }
    \label{fig:recovery}
  \vspace{-4mm}
\end{figure}

\subsection{Learned Motions in Simulation}
\label{subsec:simmotion}

Our learned policy demonstrates many interesting behaviors on "Stones-Everywhere" in simulation. In Fig.~\ref{fig:turning}, we made the robot walk around on stepping stones by providing a series of waypoints as target positions and headings, showcasing the ability of omnidirectional locomotion learned by our method. During locomotion, the robot also had different gait patterns and contact modes under different situations.

We also find the policy can recover from missed steps in simulation, as is displayed in Fig.~\ref{fig:recovery}. The $\sim 50$ kg robot is able to quickly react to the external force disturbances and continues to walk even when the forces on all feet reaches $\sim 100$ N.

\subsection{Benchmarked Ablation Studies}
\label{subsec:ablation}

We compare different settings and analyze the robustness of the learned policies on "Stones-Everywhere". The compared settings are:
\begin{enumerate}
    \item w/o navigation: where the velocity tracking formulation is used based on \cite{rudin2022learning} with the corresponding curriculum, while the intrinsic rewards and the symmetry-based data augmentation are also applied;
    \item w/o curriculum: every technique proposed in this paper except the improved curriculum is used, and the curriculum is the one in \cite{rudin2022advanced};
    \item w/o curiosity: every technique proposed in this paper except the intrinsic curiosity rewards is used;
    \item w/o augmentation: every technique proposed in this paper except the data augmentation is used;
    \item proposed: every technique proposed in this paper is used.
\end{enumerate}

\begin{figure}[t]
  \centering
    \includegraphics[width=86mm, trim = 0.1cm 0.5cm 0cm 0cm]{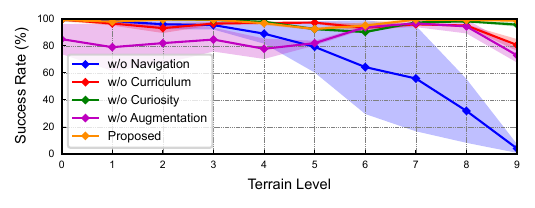}
    \caption{
       Success rate of traversing 6 meters on different levels of "Stones-Everywhere". The settings based on the navigation formulation far outperform the velocity tracking one. Three runs (policies trained with different random seeds) per setting were performed to calculate the mean (line) and std (in shadow) values, and the checkpoints of 8000 iterations (all converged) were used. The velocity command used for testing velocity tracking is $0.8$ m/s forward, which demonstrates the highest robustness empirically in simulation. Each policy's success rate is calculated from 1000 trials on 100 diverse randomized terrains.
    }
    \label{fig:complevel}
  \vspace{-3mm}
\end{figure}

\begin{figure}[t]
  \centering
    \includegraphics[width=86mm, trim = 0.1cm 0.5cm 0cm 0cm]{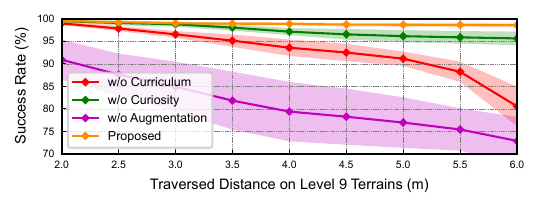}
    \caption{
       Success rate of traversing different distances on the hardest level of "Stones-Everywhere". Our proposed strategy stably learns robust policies over long distances. The shadow indicates the std values.
    }
    \label{fig:compdist}
  \vspace{-6mm}
\end{figure}

Two metrics are used for the comparison of performance and robustness. The first one is the success rate of traversing 6 meters on different difficulty levels, as is shown in Fig.~\ref{fig:complevel}. After level 4 with sparsity $\sim 40\%$, the averaged success rate for velocity tracking is lower than $90\%$ and quickly drops down as the difficulty level goes up. In comparison, the navigation formulation can achieve a high success rate, though the “w/o augmentation” setting does not perform well on low-difficulty terrains. This can be explained by the policy's inability to overcome the hardest terrains without data augmentation, and the neural network policy forgets the samples on low-difficulty terrains when the robots get stuck on high-difficulty ones.


The second metric is the success rate of traversing different distances on the highest difficulty level, which demonstrates a clearer trend between the navigation-formulated ones, as is shown in Fig.~\ref{fig:compdist}. All techniques in our strategy significantly contribute to improving robustness, with symmetry-based data augmentation being the most critical one. The improved curriculum design and the curiosity rewards have relatively smaller effects. Besides, the symmetry-based data augmentation also makes the motions more balanced with less base inclination.



\subsection{Finetuning v.s. From Scratch}

We train specialist policies on multiple challenging terrains as shown in Fig.~\ref{fig:fig1cover} and~\ref{fig:terrains} by finetuning the learned generalist policy. The generalist policy took 8000 iterations, the "Stones-2Rows" policy took 2000 iterations, the "Balance-Beams" policy took 3000 iterations, and the "Stepping-Beams" policy took 1000 iterations. In comparison, when the specialist policies are trained from scratch for 10000 iterations per task, the robot can not learn to traverse the terrains, as we design these terrains to be quite challenging.

\subsection{Transferability of Policies}

\begin{figure}[t]
  \centering
    \hspace*{-1cm}\includegraphics[width=60mm]{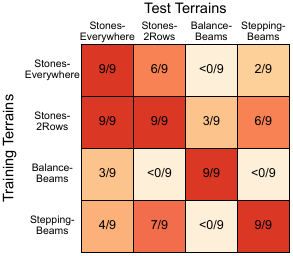}
    \caption{
       The transferability matrix of our learned policies. Values (before "/9") in the matrix are the maximum difficulty ($0\sim9$ with 9 being the hardest) of test terrains that can be traversed for $3.5\rm m$ with an $80\%$ success rate. For the "$<0$"s, the success rate is within $[50\%,80\%)$ on level $0$.
    }
    \label{fig:transfer}
  \vspace{-4mm}
\end{figure}

We also test whether policies trained on different terrains are transferable to other terrains. Figure~\ref{fig:transfer} presents the transferability matrix of our learned policies. Due to the similarity of terrains, policies trained on "Stones-Everywhere" and "Stones-2Rows" can be mutually transferred, and the latter is also transferrable to and from "Stepping-Beams". For other terrain pairs, the policies do not tranfer well. Such results indicate that the learned policies highly differentiate due to their training data although they originate from the same generalist policy. Hence, if we want a unified policy that performs well on various terrains, we must randomize all possible terrains the robot may encounter. Otherwise, the learned policy cannot generalize well.

\section{Real-World Results}

Due to practical reasons, we only build two terrains in the real world: stepping stones and balance beams, and successfully transfer the learned "Stones-2Rows" and "Balance-Beams" policies. To minimize the sim-to-real gap and focus on the locomotion problem, we use a motion capture system to get the robot's state and terrain information.


\subsection{Agile Motions}
\label{subsec:motions}

\begin{figure}[t!]
  \centering
    \includegraphics[width=86mm, trim = 0cm 0.2cm 0cm 0cm]{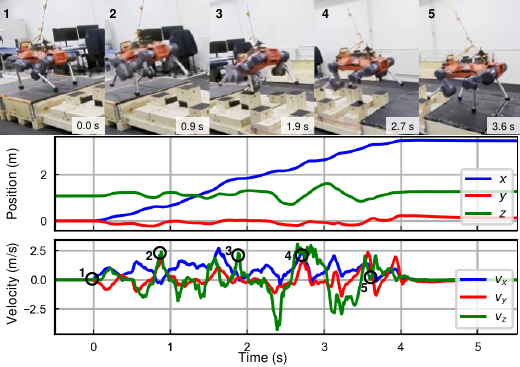}
    \caption{
       Learned agile motions on stepping stones in real-world experiments. The robot frequently leaps upwards and forwards to traverse the terrain. The robot positions are sampled from the motion capture system in the world frame, and the velocities are obtained by differentiating the positions.
    }
    \label{fig:stonemotions}
  \vspace{-2mm}
\end{figure}

\begin{figure}[t!]
  \centering
    \includegraphics[width=86mm, trim = 0cm 0.4cm 0cm 0cm]{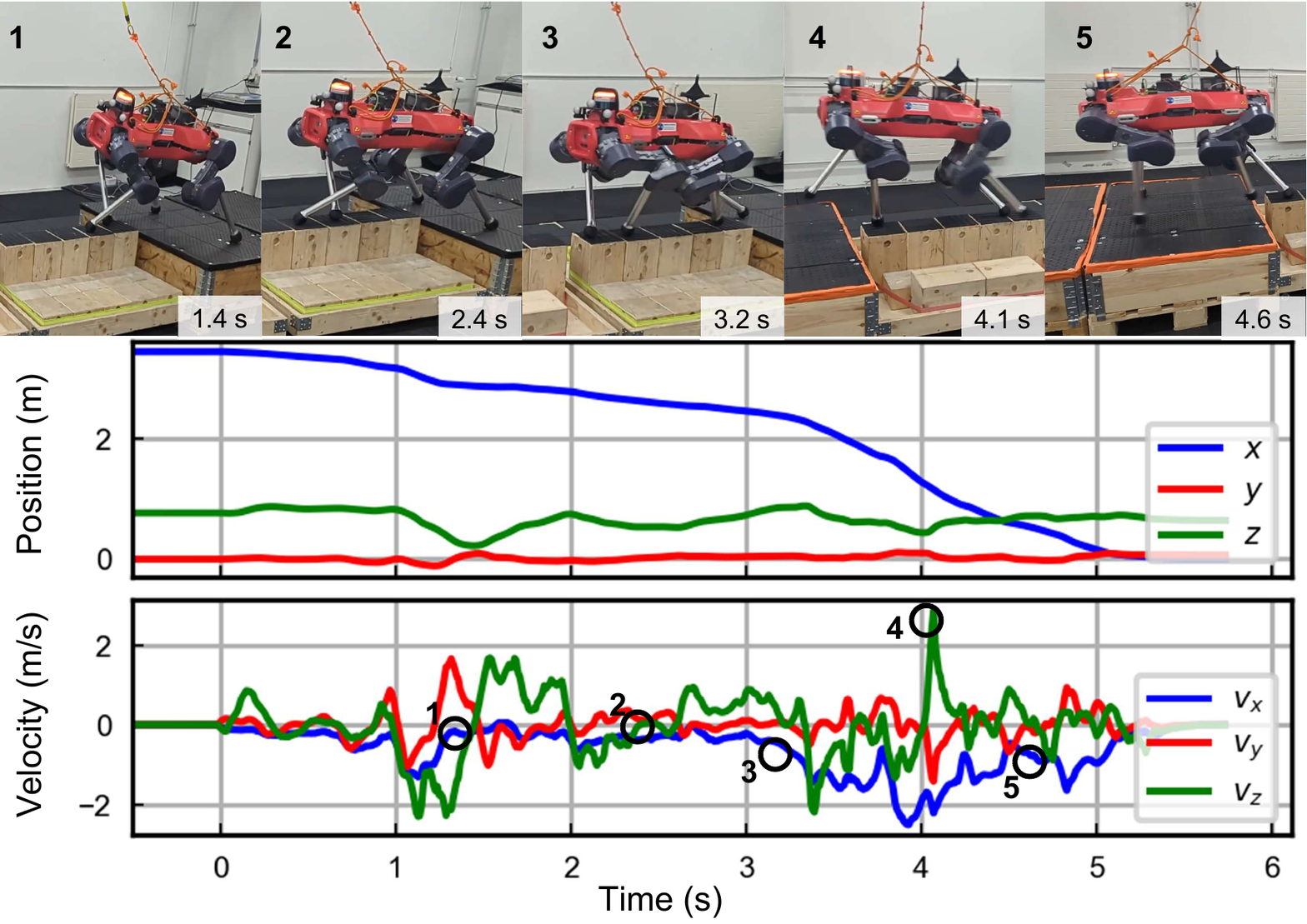}
    \caption{
       Learned agile motions on balance beams in real-world experiments. The robot first slowly aligns itself with the beam center before landing on the beam. Afterward, the robot rapidly alternates its left and right diagonals, and makes a speed burst ($3\sim 5$ s in the plot) to traverse the beams.
    }
    \label{fig:beammotions}
  \vspace{-2mm}
\end{figure}

Motions of the learned policies on real terrains are displayed in Fig. \ref{fig:stonemotions} and \ref{fig:beammotions}. The motions involve multiple leaps and rapidly changing velocities which are difficult to achieve with velocity tracking. To solve the task in the given episode length, the robot takes different strategies on different terrains. It keeps leaping on stepping stones, while first slowly aligning itself towards the center and then making a speed burst to quickly traverse the terrain on balance beams.

\subsection{Robustness Tests}
\label{subsec:robTest}

\begin{table}[t!]
\centering
\caption{Repetitive Robustness Test Results on Real Terrains}
\label{tab:sr_real}
\begin{tabular}{ccc}
\hline
\textbf{Real Terrain} &  Stones-2Rows & Balance-Beams \\ \hline
\textbf{Success Rate} & 8/10 & 5/5\\
\textbf{Survival Rate} & 10/10 & 5/5\\
\hline
\end{tabular}
\vspace{-4mm}
\end{table}

We did repetitive robustness tests for the two policies (see also the video attachment), and the results are presented in Table \ref{tab:sr_real}. The success rate counts the trials that were perfectly finished, while the survival rate also counts the cases where the robot missed a step but recovered. 

Our method also differs from the model-based controllers in that the robot can quickly react to unexpected missed steps and survive, as is discussed in Sec. \ref{subsec:simmotion}. We refer readers to the video attachment to see the emergent behaviors. It is worth mentioning that our controller is vulnerable to mapping drifts, and a 5-cm odometry drift can make the robot fall, though this is partly due to the highly risky terrains.

\section{Limitations and Discussion}

Though we can achieve agile locomotion on various terrains, we have difficulty training a unified policy for all because the learning progresses on different terrains can vary and interfere with each other. Also, we use slightly different settings for different specialist tasks with human insights to help the sim-to-real transfer. The reward functions can be tedious to tune, and the rare failure cases are hard to interpret. 

To be specific, the task reward functions are quite an "equilibrium" of multiple motivations. First, as we want the robot to automatically discover a way to reach the target without human priors, a short reward duration is preferred but can also lead to sparse rewards and hard exploration. Second, a long episode length can make the motions less aggressive for sim-to-real transfer but also makes the task rewards sparse. Hence, tuning the rewards is tedious, and we partly sidestep these trade–offs by pre-training and finetuning in this paper.

Regarding the failure cases, it is always difficult to achieve a 100\% success rate, and the rare failure cases do exist. However, contrary to model-based controllers whose failure cases are easy to understand (e.g., missed steps, slippery, or unexpected collisions), our controllers' failure cases are more difficult to interpret. In these rare cases, the robot may misplace its feet in the air when there are feasible footholds, which might be due to the singularity of neural networks \cite{orhan2018skip}. We believe representation learning techniques such as contrastive learning \cite{laskin2020curl} may help solve this.

\section{Conclusion and Future Works}

In this paper, we achieve end-to-end learning of agile quadrupedal locomotion on challenging risky terrains. To this end, we propose to use the navigation formulation and an exploration strategy with multiple improvements from previous works. We apply two-staged training to overcome exploration on challenging terrains, with specialist policies on different terrains that are finetuned from one generalist policy.

Our future work will include: 1) making the terrain perception onboard based on online representation techniques such as the one in \cite{hoeller2022neural}; 2) training a unified policy for locomotion on all kinds of terrains; and 3) making the learned policy more interpretable and robust.

\section*{APPENDIX}

\subsection{Task Specifications}

The task-related domain randomization settings are presented in Table~\ref{tab:dr}, and the used symbols and reward terms in Table~\ref{tab:reward}.

\begin{table}[H]
\centering
\caption{Domain Randomization Settings in Different Tasks}
\label{tab:dr}
\begin{tabular}{p{4.2cm}P{3.5cm}}
\hline
\textbf{Parameter}      & \textbf{Randomization Range}       \\ \hline
\textbf{Stones-Everywhere} & \\
Episode Length (s) & $[5,7]$ \\
Target Distance (m) & $[1.5,4.9]$ \\

\hline

\textbf{Stones-2Rows}   & \\
Episode Length (s) & $[6,8]$ \\
Target Distance (m) & $[3.5,4.5]$ \\
Heightmap Scan Drift (m) & $[-0.05,0.05]$ \\

\hline
\textbf{Balance-Beams}   &  \\

Episode Length (s) & $[8,10]$ \\
Target Distance (m) & $[3.5,4.5]$ \\
Heightmap Scan Drift (m) & $[-0.025,0.025]$ \\
Center of Mass Bias (m) & $x:[-0.15,0.15]$\linebreak $ y:[-0.05,0.05]$ \linebreak$ z:[-0.1,0.2]$ \\

\hline

\textbf{Stepping-Beams}   &  \\
Episode Length (s) & $[5,7]$ \\
Target Distance (m) & $[2.5,4.0]$ \\

\hline
\end{tabular}
\vspace{-6mm}
\end{table}

\begin{table}[H]
\centering
\caption{Used Symbols and Reward Function Terms}
\label{tab:reward}
\begin{tabular}{p{1.45cm}p{6.35cm}}
\hline
\textbf{Symbol}      & \textbf{Description}       \\ \hline
$t$ and $T$ & The current timestep, and the whole length of an episode \\
$p$ and $p^*$ & The 2-d position and the target position of the robot \\
$\psi$ and $\psi^*$ & The heading and the target heading of the robot \\
$q, \dot{q},$ and $\dot{q}_{\lim}$ & The joint positions, joint velocities, and the joint velocity limit of the robot \\
$v, v^H,$ and $\omega$ & The linear velocity, horizontal linear velocity, and angular velocity of the robot base \\
$z$ & The height of the robot base\\
$v_f$ and $F_f$ & The linear velocity and the contact force of the $f$-th foot \\
$a$ & The action, i.e., the target joint positions \\
$\tau$ and $\tau_{\lim}$ & The torques and the torque limit of the joints \\
$g$ & The projected gravity on the robot base \\
$\delta_T(t_0)$ & $\frac{1}{t_0}\cdot \mathds{1}(t>T-t_0)$, the reward duration mask for the last $t_0$ seconds \\
$\delta_p(d_0)$ & $\mathds{1}(\lVert{p_t}-{p_t}^*\rVert < d_0)$, the robot position mask\\
$\delta_\psi(\theta_0)$ & $\mathds{1}(\lVert \psi_t-\psi_t^*\rVert<\theta_0)$, the robot heading mask \\
"S2" & The abbreviation for "Stones-2Rows"  \\
"BB" & The abbreviation for "Balance-Beams"\\
\hline\\
\end{tabular}

\resizebox{86mm}{!}{%
\begin{tabular}{p{2.5cm} p{3.6cm} p{2.9cm}}
\hline
\textbf{Reward Term}      & \textbf{Expression}  & \textbf{Weight}     \\ \hline
Position Tracking &$ \frac{1}{1+\left\|{p_t}-{p_t}^*\right\|^2} $  & $10 \delta_T(2)$ \\ 
 & \hfill  \textit{for "S2" and "BB"} & $25\delta_T(4)$ \\[1mm]
Heading Tracking &$ \frac{1}{1+\left\|{\psi_t}-{\psi_t}^*\right\|^2} $  & $5 \delta_p(2) \delta_T(4)$ \\ 
 & \hfill  \textit{for "S2" and "BB"} & $12\delta_p(2.5)\delta_T(4)$ \\[1mm]
Termination Penalty    & const for early termination & $-200$\\ 
Collision Penalty  & const for thighs and shanks & $-1$ \\ 
Joint Velocity Penalty &$\lVert \dot{q}_t \rVert ^2$ & $-0.001$ \\ 
Joint Velocity Limit &$\sum_{i=1}^{12} \max (\lvert\dot{q}_{i,t}\rvert-0.9\dot{q}_{\lim},0)$ & $-1 $ \\ 
Base Accel. Penalty   &$\lVert \dot{v}_t \rVert ^2 + 0.02\lVert \dot{\omega}_t \rVert ^2$ & $-0.001$ \\ 
Feet Accel. Penalty   &$ \sum_{f=1}^4 \lVert \dot{v}_{f,t}\rVert$ & $-0.0005$ \\ 
Action Rate Penalty   &$\lVert a_t-a_{t-1} \Vert ^2$ & $-0.01$ \\ 
Torque Penalty   &$  \lVert \tau_t\rVert^2$ & $-0.00001$\\ 
 & \hfill  \textit{for "BB"} & $-0.00002$ \\[1mm]
Torque Limit & $ \sum_{i=1}^{12} \max (\lvert\tau_{i,t}\rvert-\tau_{\lim},0)$ & $-0.2$ \\ 
 & \hfill  \textit{ for "BB" } w/ $\tau_{\lim}' = 0.8\tau_{\lim}$ & $-0.5$ \\[1mm]
Contact Force Penalty  & $\sum_{f=1}^4 \text{clip}(\lVert F_{f,t} \rVert - 700, 0,700)^2$ & $-0.000025$ \\ 
Don't Wait & $ \mathds{1}(\lVert v_t \rVert < 0.2)$ & $-(1-\delta_p(1))$ \\ 
Move in Direction & $ \cos \langle v_t, p_t^*-p_t \rangle$  & $1$ for first 150 iter\\ 
Stand Still & $ 2.5\lVert v_t \rVert +1\lVert \omega_t \rVert$ & $- \delta_p(0.25)  \delta_\psi(0.5) \delta_T(1)$\\ 
Aggressive Motion & $(\lvert v_t^H \rvert - 1) ^2\cdot \mathds{1}(\lvert v_t^H \rvert>1)$ & 0\\
 & \hfill  \textit{for "S2" and "BB"} & $-5$ \\[1mm]
Stand Pose &  $ \lvert z_t -0.6 \rvert + g_{x,t}^2+g_{y_t}^2 $ & $0$ \\
 & \hfill  \textit{for "S2" and "BB"} & $-5\delta_p(0.25)  \delta_\psi(0.5) \delta_T(1)$ \\
\hline
\end{tabular}}
\vspace{-3mm}
\end{table}


\subsection{Learning Settings}

Some learning-related settings are presented in Table~\ref{tab:rlconfig}.

\begin{table}[H]
\centering
\caption{Learning Settings}
\label{tab:rlconfig}
\begin{tabular}{p{3.1cm}P{4.5cm}}
\hline
\textbf{Configuration}      & \textbf{Values}       \\ \hline

Actor and critic & MLP, hidden layer size $(512, 256, 128)$\\
 & All observations flattened \\
Exteroception & $25\times 16$ heightmap, $7$cm gridsize \\ 
Number of robots & $4096$ \\
Policy steps per iteration & $48$ \\
Clip negative rewards to 0? & No\\
RND M1 & MLP, hidden layer size $(128, 128)$\\
RND M2 & MLP, hidden layer size $(256, 256)$\\
RND reward weight & $1.0$ \\
RND optimization weight & $1.0$ \\
Loss for RND M1 & mean L1-loss\\

\hline

\end{tabular}
\end{table}


\section*{ACKNOWLEDGMENT}

This work is supported in part by the EU Horizon 2020 program grant
agreement No.852044

\bibliographystyle{IEEEtran}
\balance
\bibliography{IEEEabrv,Reference}

\end{document}